\def\year{2019}\relax
\documentclass[letterpaper]{article} %DO NOT CHANGE THIS
\usepackage{aaai19}  %Required
\usepackage{times}  %Required
\usepackage{helvet}  %Required
\usepackage{courier}  %Required
\usepackage{url}  %Required
\usepackage{graphicx}  %Required
\usepackage{array}
\usepackage{amsmath}
\usepackage{amssymb}
\usepackage{times}
\usepackage{textcomp}
\usepackage{amsfonts}

\usepackage{algorithm}
\usepackage{algorithmicx}
\usepackage[noend]{algpseudocode}
\usepackage{setspace}

\usepackage{CJK}
\usepackage{color}
\usepackage{multicol}
\usepackage{multirow}
\usepackage{array}
\usepackage{balance}
\usepackage{arydshln}
\usepackage{tikz}
\usepackage{url}

\DeclareSymbolFont{extraup}{U}{zavm}{m}{n}
\DeclareMathSymbol{\varheart}{\mathalpha}{extraup}{86}
\DeclareMathSymbol{\vardiamond}{\mathalpha}{extraup}{87}

\usepackage{xspace}

\usepackage{helvet}
\usepackage{courier}

\usepackage{bm}
\usepackage{color}

\usepackage{arydshln}

\usepackage{amsfonts}
\usepackage{graphicx} 
\usepackage{graphics}
\usepackage{standalone}

\usepackage{caption}
\usepackage{subfig}
\usepackage{balance}

\usepackage{tikz}
\usetikzlibrary{arrows.meta,calc,decorations.markings,math,arrows.meta}
\usetikzlibrary{decorations.pathreplacing}

\makeatletter
\newcommand{\@BIBLABEL}{\@emptybiblabel}
\newcommand{\@emptybiblabel}[1]{}
\makeatother
\usepackage[hidelinks]{hyperref}
\title{Dynamic Layer Aggregation for Neural Machine Translation \\ with Routing-by-Agreement}

\author{Zi-Yi Dou\\\normalsize Carnegie Mellon University\\{\normalsize \tt zdou@andrew.cmu.edu} \And
Zhaopeng Tu\thanks{Zhaopeng Tu is the corresponding author. Work was done when Zi-Yi Dou was interning at Tencent AI Lab.}\\\normalsize Tencent AI Lab\\{\normalsize \tt zptu@tencent.com} \And
Xing Wang\\\normalsize Tencent AI Lab\\{\normalsize \tt brightxwang@tencent.com} \AND
Longyue Wang\\\normalsize Tencent AI Lab\\{\normalsize \tt vinnylywang@tencent.com} \And
Shuming Shi\\\normalsize Tencent AI Lab\\{\normalsize \tt shumingshi@tencent.com} \And
Tong Zhang\\\normalsize Tencent AI Lab\\{\normalsize \tt bradymzhang@tencent.com}
}

\begin{document}
\maketitle

\begin{abstract}

%%%% Version 1 (ziyi->xing) %%%%%
% Deep neural networks have been widely used in various applications and achieved exciting performance among tasks. Fusing information across layers has already proven to be useful in computer vision community and recent studies have begun to explore aggregating layers in the field of neural machine translation (NMT). However, most of the previous methods combine layers in a static fashion in that their aggregation strategy is independent of specific hidden states. We argue that treating each combinations of layers individually is a necessary and important step. Inspired by routing-by-agreement in {\color{red}capsule} networks, in this work we propose several ways to aggregate layers dynamically, meaning we could learn specific weights of combination for given hidden states.  {\color{red}To the best of our knowledge, the proposed method is the first to explore capsule networks in NMT.} We implement our method upon the state-of-the-art NMT system, namely the Transformer. Experimental results on widely-used WMT14 English$\Rightarrow$ German and WMT17 Chinese$\Rightarrow$ English translation data demonstrate the effectiveness and universality of the proposed method. 

%%%% Version 2 (ziyi->xing->longyue) %%%%%
With the promising progress of deep neural networks, layer aggregation has been used to fuse information across layers in various fields, such as computer vision and machine translation. However, most of the previous methods combine layers in a static fashion in that their aggregation strategy is independent of specific hidden states. 
{Inspired by recent progress on capsule networks, in this paper we propose to use {\em routing-by-agreement} strategies to aggregate layers dynamically. Specifically, the algorithm learns the probability of a part (individual layer representations) assigned to a whole (aggregated representations) in an iterative way and combines parts accordingly. }
We implement our algorithm on top of the state-of-the-art neural machine translation model {\textsc{Transformer}} and conduct experiments on the widely-used WMT14 English$\Rightarrow$German and WMT17 Chinese$\Rightarrow$English translation datasets. Experimental results across language pairs show that the proposed approach consistently outperforms the strong baseline model and 
a representative static aggregation model.

%建议 先定性质 后给实验结论 这一句前提了 wangxing
%Also, our method is the first to explore capsule networks in NMT to our knowledge. 
\end{abstract}

\section{Introduction}
Deep neural networks have advanced the state of the art in various communities, from computer vision to natural language processing. Researchers have directed their efforts into designing patterns of modules that can be assembled systematically, which makes neural networks deeper and wider. However, one key challenge of training such huge networks lies in how to transform and combine information across layers. 
%However, training networks consisting of these amounts of layers is extremely difficult and gradient vanishing/exploding would be a huge problem. 
To encourage gradient flow and feature propagation, researchers in the field of computer vision have proposed various approaches
%to combine information across different layers
, such as residual connections~\cite{he2016deep}, densely connected network~\cite{Huang:2017:CVPR} and deep layer aggregation~\cite{Yu:2018:CVPR}. 

State-of-the-art neural machine translation (NMT) models generally implement encoder and decoder as multiple layers~\cite{wu2016google,Gehring:2017:ICML,Vaswani:2017:NIPS,chen2018the}, in which only the top layer is exploited in the subsequent processes.
Fusing information across layers for deep NMT models, however, has received substantially less attention. 
A few recent studies reveal that simultaneously exposing all layer representations outperforms methods that utilize just the top layer for natural language processing tasks~\cite{Peters:2018:NAACL,shen2018dense,wang2018multi,Dou:2018:EMNLP}.
However, their methods mainly focus on {\em static} aggregation in that the aggregation mechanisms are the same across different positions in the sequence. Consequently, useful context of sequences embedded in the layer representations are ignored, which could be used to further improve layer aggregation.

In this work, we propose dynamic layer aggregation approaches, which allow the model to aggregate hidden states across layers for each position dynamically. We assign a distinct aggregation strategy for each symbol in the sequence, based on the corresponding hidden states that represent both syntax and semantic information of this symbol. To this end, we propose several strategies to model the dynamic principles. First, we propose a simple {\em dynamic combination} mechanism, which assigns a distinct set of aggregation weights, learned by a feed-forward network, to each position. Second, inspired by the recent success of {\em iterative routing} on assigning parts to wholes for computer vision tasks~\cite{Sabour:2017:NIPS,Hinton:2018:ICLR}, here we apply the idea of {\em routing-by-agreement} to layer aggregation. Benefiting from the high-dimensional coincidence filtering, {\it i.e.} the agreement between every two internal neurons, the routing algorithm has the ability to extract the most active features shared by multiple layer representations.

We evaluated our approaches upon the standard \textsc{Transformer} model~\cite{Vaswani:2017:NIPS} on two widely-used WMT14 English$\Rightarrow$German and WMT17 Chinese$\Rightarrow$English translation tasks. 
% We employed \textsc{Transformer}~\cite{Vaswani:2017:NIPS} as the baseline system since it has proven to outperform other architectures on the two tasks~\cite{Vaswani:2017:NIPS,hassan2018achieving}.
%{Experimental results demonstrate that one of the representative static layer aggregation} strategies -- linear combination indeed improves translation performance, indicating the necessity 
{We show that although static layer aggregation} strategy indeed improves translation performance, which indicates the necessity 
and {effectiveness} 
of fusing information across layers for deep NMT models,
our proposed dynamic approaches outperform their static counterpart. Also, our models consistently improve translation performance over the vanilla \textsc{Transformer} model across language pairs.
It is worth mentioning that \textsc{Transformer-Base} with dynamic layer aggregation outperforms the vanilla \textsc{Transformer-Big} model with only less than half of the parameters.

\paragraph{\bf Contributions.} Our key contributions are:
\begin{itemize}
    \item Our study demonstrates the necessity and {effectiveness} of dynamic layer aggregation for NMT models, which benefits from exploiting useful context embedded in the layer representations. 
    \item {Our work is among the few studies ({\it cf.} ~\cite{Gong:2018:arXiv,zhao2018investigating}) which prove that the idea of capsule networks can have promising applications on natural language processing tasks.}
\end{itemize}

\section{Background}

\subsection{Deep Neural Machine Translation}

Deep representations have a noticeable effect on neural machine translation~\cite{Meng:2016:ICLRWorkshop,Zhou:2016:TACL,wu2016google}.
Generally, multiple-layer encoder and decoder are employed to perform the translation task through a series of nonlinear transformations from the representation of input sequences to final output sequences.

Specifically, the encoder is composed of a stack of $L$ identical layers with the bottom layer being the word embedding layer. Each encoder layer is calculated as
\begin{eqnarray}
    {\bf H}_e^l = \textsc{Layer}_e({\bf H}_e^{l-1}) + {\bf H}_e^{l-1}
\end{eqnarray}
where a residual connection~\cite{he2016deep} is employed around each of the two layers. $\textsc{Layer}(\cdot)$ is the layer function, which can be implemented as RNN~\cite{cho2014learning}, CNN~\cite{Gehring:2017:ICML}, or self-attention network (SAN)~\cite{Vaswani:2017:NIPS}. In this work, we evaluate the proposed approach on the standard Transformer model, while it is generally applicable to any other type of NMT architectures.

The decoder is also composed of a stack of $L$ layers:
\begin{eqnarray}
    {\bf H}_d^l = \textsc{Layer}_d({\bf H}_d^{l-1}, {\bf H}_e^L) + {\bf H}_d^{l-1}
\end{eqnarray}
which is calculated based on both the lower decoder layer and the top encoder layer ${\bf H}^{L}_{e}$. The top layer of the decoder ${\bf H}_d^L$ is used to generate the final output sequence.

%%%\vspace{5pt}

As seen, both the encoder and decoder stack layers in sequence and only utilize the information in the top layer. 
While studies have shown deeper layers extract more semantic and more global features~\cite{zeiler2014visualizing,Peters:2018:NAACL}, these do not prove that the last layer is the ultimate representation for any task. 
Although residual connections have been incorporated to combine layers, these connections have been ``shallow'' themselves, and only fuse by simple, one-step operations~\cite{Yu:2018:CVPR}.

\subsection{Exploiting Deep Representations}

Recently, aggregating layers to better fuse semantic and spatial information has proven to be of profound value in computer vision tasks~\cite{Huang:2017:CVPR,Yu:2018:CVPR}.
% In natural language processing community,~\citeauthor{Peters:2018:NAACL}~\shortcite{Peters:2018:NAACL} and~\citeauthor{shen2018dense}~\shortcite{shen2018dense} have proven that simultaneously exposing all layer representations outperforms methods that utilize just the top layer on several generation tasks.
{For machine translation,~\citeauthor{shen2018dense}~\shortcite{shen2018dense} and \citeauthor{Dou:2018:EMNLP}\shortcite{Dou:2018:EMNLP} have proven that simultaneously exposing all layer representations outperforms methods that utilize just the top layer on several generation tasks.
Specifically, one of the methods proposed by \citeauthor{Dou:2018:EMNLP}\shortcite{Dou:2018:EMNLP} is to linearly combine the outputs of all layers:}
\begin{equation}
    \mathbf{\widetilde{H}} = \sum_{l=1}^L {\bf W}_l \mathbf{H}^l      
    \label{eqn:linear}
\end{equation}
where $\{{\bf W}_1, \dots, {\bf W}_L\} \in \mathbb{R}^{d}$ are trainable parameter matrices, where $d$ is the dimensionality of hidden layers. The linear combination strategy is applied to both the encoder and decoder. The combined layer $\mathbf{\widetilde{H}}$ that embeds all layer representations instead of only the top layer ${\bf H}^L$, is used in the subsequent processes.

%%%\vspace{5pt}
As seen, the linear combination is encoded in a static set of weights $\{{\bf W}_1, \dots, {\bf W}_L\}$, which ignores the useful context of sentences that could further improve layer aggregation. In this work, we introduce the dynamic principles into layer aggregation mechanisms.

\section{Approach}

\iffalse

\subsection{Static Combination}

First, we investigate a simple static method to linearly combine the outputs of all layers:
\begin{equation}
    \mathbf{\widehat{H}} = \sum_{l=1}^L {\bf W}_l \mathbf{H}^l,      
    \label{eqn:linear}
\end{equation}
where $\{{\bf W}_1, \dots, {\bf W}_L\} \in \mathbb{R}^{d}$ are trainable parameter matrices, where $d$ is the dimensionality of hidden layers. The linear combination strategy is applied to both the encoder and decoder. The combined layer $\mathbf{\hat{H}}$ that embeds all layer representations instead of only the top layer ${\bf H}^L$, is used in the subsequent processes.

%%%\vspace{5pt}
As seen, the linear combination is encoded in a static set of weights $\{{\bf W}_1, \dots, {\bf W}_L\}$, which ignores the useful context of sentences that could further improve layer aggregation. In the following part, we propose dynamic layer aggregation mechanisms, which assigns weights to different combinations of layers dynamically. {\color{red} (needs to be polished, one sentence to summarize the basic idea of our work.)}

\fi

\subsection{Dynamic Combination}

An intuitive extension of static linear combination is to generate different weights for each layer combination rather than apply the same weights all the time.
To this end, we calculate the weights of the linear combination as
\begin{align}
{\bf W}_l = \textsc{Ffn}_l({\bf H}^1, \dots, {\bf H}^L)         &&    \in \mathbb{R}^{J \times d}
\end{align}
where $J$ is the length of the hidden layer ${\bf H}^l$, and $\textsc{Ffn}_l(\cdot)$ is a distinct feed-forward network associated with the $l$-th layer ${\bf H}^l$.
Specifically, we use all the layer representations as the context, based on which we output a weight matrix that shares the same dimensionality with ${\bf H}^l$.
Accordingly, the weights are adapted during inference depending on the input layer combination.

Our approach has two strengths. First, it is a more flexible strategy to dynamically combine layers by capturing contextual information among them, which is ignored by the conventional version. 
Second, the transformation matrix $\textsc{Ffn}_l(\cdot)$ offers the ability to assign a distinct weight to each state in the layers, while its static counterpart fails to exploit such strength since the length of input layers $J$ varies across sentences thus cannot be pre-defined.

% , which reads all the layer representations and outputs a weight matrix that share the same dimensionality with ${\bf H}^l$.
% While the conventional linear combination only allows a set of static weights shared by all sentences, we propose a more flexible strategy to compute dynamic weights based on the specific sentence context.
% Accordingly, our strategy is able to find an effective way to combine layers by capturing useful contextual information among them, which is ignored by the conventional version.

% Instead of context-agnostic weights in the conventional linear combination, we assign a distinct weight to each state in the hidden layer. 
% In this way, our model is capable of recognizing useful part of the hidden states of different layers and finding an effective way to combine them. On the contrary, the conventional strategy only statically aggregates the information in each layer and ignore the specific information each layer captures, which would be suboptimal and lose a great amount of information.

% \vspace{10pt}

\subsection{Layer Aggregation as Capsule Routing}

The goal of layer aggregation is to find a whole representation of the input from partial representations captured by different layers.
This is identical to the aims of capsule network, which becomes an appealing alternative to solving the problem of {\em assigning parts to wholes}~\cite{hinton2011transforming}. Capsule network employs a fast iterative process called {\em routing-by-agreement}. Concretely, the basic idea is to iteratively update the proportion of how much a part should be assigned to a whole, based on the agreement between parts and wholes. 
An important difference between iterative routing and layer aggregation is that the former provides a new way to aggregate information according to the representation of the final output.
%An important difference between capsules and standard neural nets is that the activation of a capsule is calculated by a comparison between multiple incoming votes, while in a standard neural net it is based on a comparison between a single input vector and a learned weight vector.

% The activities of the neurons within an active capsule represent the various properties of a particular entity that is present in the image. These properties can include many different types of instantiation parameter such as pose (position, size, orientation), deformation, velocity, albedo, hue, texture, etc.

A capsule is a group of neurons whose outputs represent different properties of the same entity from the input~\cite{Hinton:2018:ICLR}. Similarly, a layer consists of a group of hidden states that represent different linguistic properties of the same input
%, such as both local and global syntax, and context-dependent aspects of word meaning
%Given the layer representations as input, these properties can include many different linguistic properties such as both local and global syntax, and context-dependent aspects of word meaning
% We expect capsule networks can help to extract different linguistic properties from layer representations. 
~\cite{Peters:2018:NAACL,Anastasopoulos:2018:NAACL}, thus each hidden layer can be viewed as a capsule. 
Given the layers as input capsules, we introduce an additional layer of output capsules and then perform iterative routing between these two layers of capsules.
Specifically, in this work we explore two representative routing mechanisms, namely {\em dynamic routing} and {\em EM routing}, which differ at how the iterative routing procedure is implemented.
We expect layer aggregation can benefit greatly from advanced routing algorithms, which allow the model to
allow the model to directly learn the part-whole relationships.
% exploit a much more complicated non-linearity than the standard neural nets.

\subsubsection{Dynamic Routing}

\begin{figure}[t]
\centering
\includegraphics[width=0.4\textwidth]{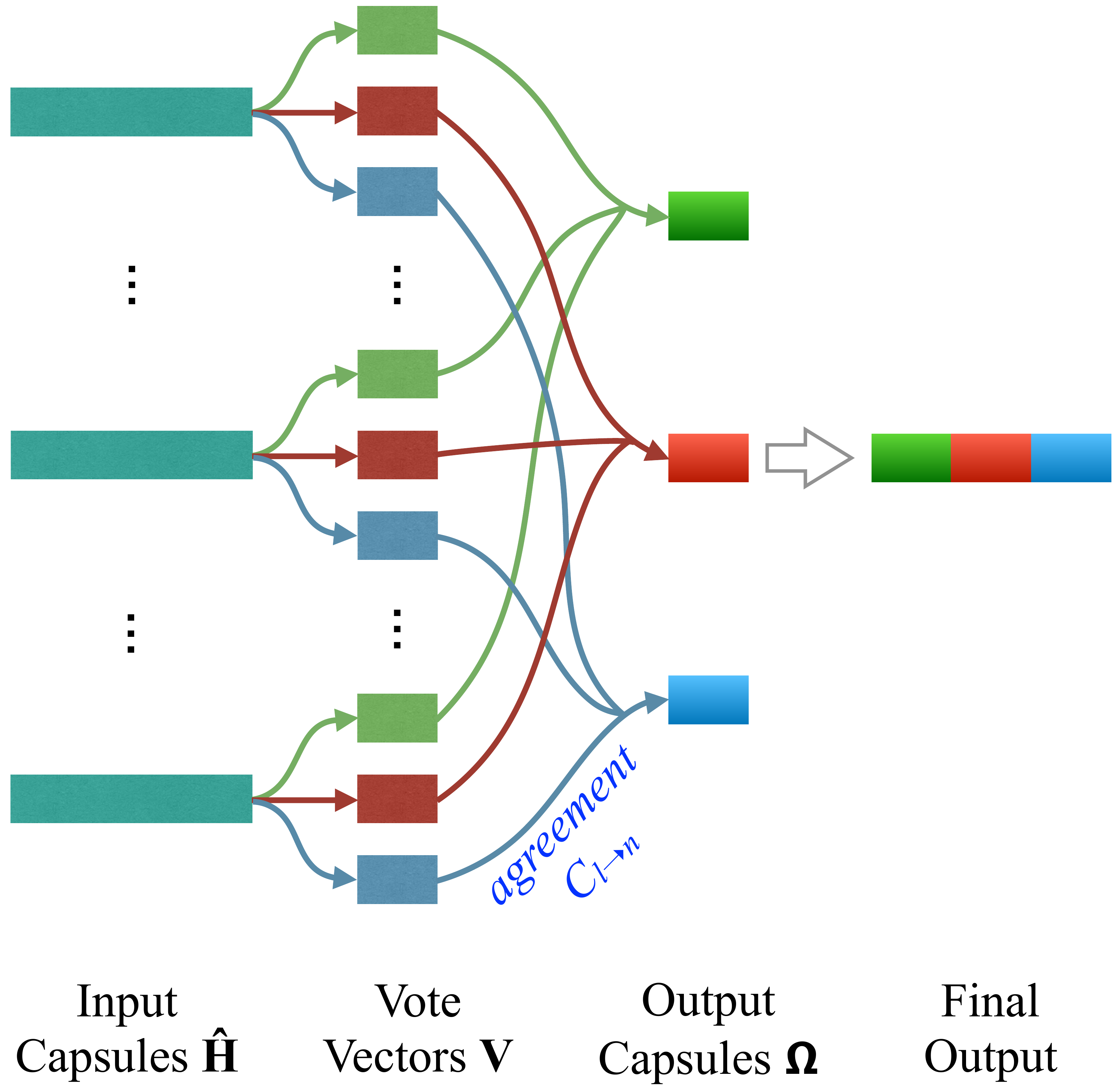}
\caption{Illustration of the dynamic routing algorithm.}
\label{fig:dynamic}
\end{figure}

Dynamic routing is a straightforward implementation of routing-by-agreement. To illustrate, the information of $L$ input capsules is dynamically routed to $N$ output capsules, which are concatenated to form the final output ${\bf \widetilde{H}}=[{\bf \Omega}_1, \dots, {\bf  \Omega}_N]$, %{\color{red} Dou:  Concatenation is not part of the dynamic routing. Actually it could be considered as a novelty (But it also has been proposed in "Information Aggregation via Dynamic Routing for Sequence Encoding".)}, 
as shown in Figure~\ref{fig:dynamic}.
Each vector output of capsule $n$ is calculated with a non-linear ``squashing'' function~\cite{Sabour:2017:NIPS}: 
\begin{eqnarray}
    {\bf  \Omega}_n &=& \frac{||\mathbf{S}_n||^2}{1 + ||\mathbf{S}_n||^2}  \frac{\mathbf{S}_n}{||\mathbf{S}_n||}     \label{eqn:capsule}\\
    {\bf S}_n &=& \sum_{l=1}^L C_{l \rightarrow n} V_{l \rightarrow n}  
\end{eqnarray}
where $\mathbf{S}_n$ is the total input of capsule ${\bf \Omega}_n$, which is a weighted sum over all ``vote vectors'' $V_{* \rightarrow n}$ transformed from the input capsules ${\bf \widehat{H}}$:
\begin{eqnarray}
    \label{dou}
    V_{l \rightarrow n} &=& {\bf W}_{l \rightarrow n} {\bf \widehat{H}}^l
    %\textsc{Ffn}_{l \rightarrow n} ({\bf H}^1, \dots, {\bf H}^L)
\end{eqnarray}
where ${\bf W}_{l \rightarrow n}(\cdot)$ is a trainable transformation matrix, and ${\bf \widehat{H}}^l$ is an input capsule associated with input layer ${\bf H}^l$:
\begin{equation}
{\bf \widehat{H}}^l = F_l({\bf H}^1, \dots, {\bf H}^L)
\label{eqn:input-capsule}
\end{equation}
where $F_l(\cdot)$ is a distinct transformation function.\footnote{Note that we calculate each input capsule with $F_l ({\bf H}^1, \dots, {\bf H}^L)$ instead of $F_l({\bf H}^l)$, since the former achieves better performance on translation task by exploiting more context as shown in our experiment section.}
% distinct feed-forward network with \textsc{Relu} activation, which is associated with transforming the input layer ${\bf H}^l$ for the capsule ${\bf  \Omega}_n$.\footnote{Note that different from~\cite{Sabour:2017:NIPS}, we calculate the vote vector $V_{l \rightarrow n}$ with $\textsc{Ffn}_{l \rightarrow n} ({\bf H}^1, \dots, {\bf H}^L)$ instead of ${\bf W}_{l \rightarrow n} {\bf H}^l$, since we found in our preliminary experiments that the former achieves better performance on translation task by exploiting more context and non-linearities. Specifically, we have $L \times N$ distinct $\textsc{Ffn}_{l \rightarrow n}(\cdot)$ networks for each ($l$, $n$) pair.}
$C_{l \rightarrow n}$ is the assignment probability (i.e. {\em agreement}) that is determined by the iterative dynamic routing.

% {\color{red} Dou: Eqn.~\ref{dou} could be thought of as we first do a nonlinear transformation $g$ to extract feature so that ${\bf H}^1_{new}, \dots, {\bf H}^L_{new} = g({\bf H}^1, \dots, {\bf H}^L) $. Then we could write $V_{l \rightarrow n}={\bf W}_{l \rightarrow n} {\bf H}^l_{new}$.}

\begin{algorithm}[t]
\setstretch{1.2}
\caption{\label{alg:dynamic} Iterative Dynamic Routing. Input: input capsules ${\bf \widehat{H}}=\{{\bf \widehat{H}}^1, \dots, {\bf \widehat{H}}^L\}$, iterations $T$; Output: capsules ${\bf \Omega} = \{{\bf \Omega}_1, \dots, {\bf \Omega}_N\}$.}
\begin{algorithmic}[1]
\Procedure{Routing}{${\bf \widehat{H}}$, $T$}:
	\State{$\forall (\mathbf{\widehat{H}}^l, \mathbf{\Omega}_n)$: $B_{l \rightarrow n} = 0$}
	\For{$T$ iterations}
		\State{$\forall (\mathbf{\widehat{H}}^l, \mathbf{\Omega}_n)$:\ \ $C_{l \rightarrow n} = \text{softmax}(B_{l \rightarrow n})$  }
		\State{$\forall \mathbf{\Omega}_n$:\ \ compute $\mathbf{\Omega}_{n}$ by Eq.~\ref{eqn:capsule}}
		\State{$\forall (\mathbf{\widehat{H}}^l, \mathbf{\Omega}_n)$: $B_{l \rightarrow n} \mathrel{+}= \mathbf{\Omega}_n \cdot V_{l \rightarrow n}$}
	\EndFor
	\Return $\bf \Omega$
\EndProcedure
\end{algorithmic}
\end{algorithm}

Algorithm~\ref{alg:dynamic} lists the algorithm of iterative dynamic routing. The assignment probabilities associated with each input capsule ${\bf \widehat{H}}^l$ sum to 1: $\sum_n C_{l \rightarrow n} = 1$, and are determined by a ``routing softmax'' (Line 4):
\begin{equation}
C_{l \rightarrow n} = \frac{\exp(B_{l \rightarrow n})}{\sum_{n'=1}^N\exp(B_{l \rightarrow n'}) }
\end{equation}
where $B_{l \rightarrow n}$ measures the degree that ${\bf \widehat{H}}^{l}$ should be coupled to capsule $n$ (similar to energy function in the attention model~\cite{Bahdanau:2015:ICLR}), which is initialized as all 0 (Line 2).
The initial assignment probabilities are then iteratively refined by measuring the agreement between the vote vector $V_{l \rightarrow n}$ and capsule $n$ (Lines 4-6), which is implemented as a simple scalar product $\alpha_{l \rightarrow n} = \mathbf{\Omega}_n \cdot V_{l \rightarrow n}$ in this work (Line 5).

With the iterative routing-by-agreement mechanism, an input capsule %${\bf \widehat{H}}^l$ 
prefers to send its representation to output capsules, whose activity vectors have a big scalar product with the vote $V$ coming from the input capsule. Benefiting from the high-dimensional coincidence filtering, capsule neurons are able to ignore all but the most active feature from the input capsules.
Ideally, each capsule output represents a distinct property of the input.
To make the dimensionality of the final output be consistent with that of hidden layer (i.e. $d$), the dimensionality of each capsule output is set to $d/N$. 
% To make the dimensionality of the final output be consistent with hidden layer, we use $d$ capsules (i.e. $N = d$) with a scalar output. 

% Although the basic idea of iterative-updating routing is similar to dynamic linearly combination, the routing algorithm is more sophisticated and its successful applications in many fields suggest iterative-updating routing algorithm is worth being investigated. {\color{red} needs to be polished, maybe more detailed advantages supported by the experimental findings.}

\subsubsection{EM Routing}

\begin{algorithm}[t]
\setstretch{1.2}
\caption{\label{alg:em} Iterative EM Routing returns {\bf activation} $A^{\Omega}$ of the output capsules, given the {\bf activation} $A^{H}$ and {\bf vote} $V$ of the input capsule.} % $V^{i}_{l \rightarrow n}$ is the $i^{th}$ dimension of the vote from input layer ${\bf H}^{l}$ with activation $A^{H}_l$.}
\begin{algorithmic}[1]

\Procedure{EM Routing}{$A^H, V$}:
    \State{$\forall (\mathbf{\widehat{H}}^l, \mathbf{\Omega}_n)$: $C_{l \rightarrow n} = 1/N$}
	\For{$T$ iterations}
	    \State{$\forall \mathbf{\Omega}_n$:\ \ M-Step($C, A^{H}, V$)}
	    \State{$\forall \mathbf{\widehat{H}}^l$: E-Step($\mu, \sigma, A^{\Omega}, V$)}
	\EndFor
	\State{$\forall {\bf \Omega}_n$: ${\bf \Omega}_n = A^{\Omega}_n * \mu_n$}
	
	\Return ${\bf \Omega}$
\EndProcedure
\end{algorithmic}

\begin{algorithmic}[1]
\Procedure{M-Step}{$C, A^{H}, V$}\\     \Comment{hold $C$ constant, adjust ($\mu_n, \sigma_n, A^{\Omega}_n$) for $\mathbf{\Omega}_n$}%for output ${\bf \Omega}_n$}

\State{$\forall {\bf \widehat{H}}^l$: $C_{l \rightarrow n} = C_{l \rightarrow n} * A^{H}_l$ }

\State{Compute $\mu_n, \sigma_n$ by Eq.~\ref{eqn:mu} and~\ref{eqn:sigma}}

\State{Compute $A^{\Omega}_n$ by Eq.~\ref{eqn:activation}}
\EndProcedure
\end{algorithmic}

\begin{algorithmic}[1]
\Procedure{E-Step}{$\mu, \sigma, A^{\Omega}, V$}\\    \Comment{hold ($\mu, \sigma, A^{\Omega}$) constant, adjust $C_{l \rightarrow *}$ for ${\bf \widehat{H}}^l$} %for input ${\bf H}^l$}
\State{$\forall {\bf \Omega}_n$: compute $C_{l \rightarrow n}$ by Eq.~\ref{eqn:assignment}}
\EndProcedure
\end{algorithmic}
\end{algorithm}

Dynamic routing uses the cosine of the angle between two vectors to measure their agreement: $\mathbf{\Omega}_n \cdot V_{l \rightarrow n}$. The cosine saturates at 1, which makes it insensitive to the difference between a quite good agreement and a very good agreement. In response to this problem,~\citeauthor{Hinton:2018:ICLR}~\shortcite{Hinton:2018:ICLR} propose a novel Expectation-Maximization routing algorithm.
% {\color{red} polish the advantages of EM routing}

Specifically, the routing process fits a mixture of Gaussians using Expectation-Maximization (EM) algorithm, where the output capsules play the role of Gaussians and the means of the activated input capsules play the role of the datapoints. It iteratively adjusts the means, variances, and activation probabilities of the output capsules, as well as the assignment probabilities $C$ of the input capsules, as listed in Algorithm~\ref{alg:em}.
Comparing with the dynamic routing described above, the EM routing assigns means, variances, and activation probabilities for each capsule, which are used to better estimate the agreement for routing.

The activation probability  $A^H_l$ of the input capsule ${\bf \widehat{H}}^l$ is calculated by
\begin{equation}
   A^H_l = {\bf W}^H_l {\bf \widehat{H}}^l  % \textsc{Ffn}_l ({\bf H}^l)
\end{equation}
where ${\bf W}^H_l$ is a trainable transformation matrix, and ${\bf \widehat{H}}^l$ is calculated by Equation~\ref{eqn:input-capsule}. The activation probabilities $A^H$ and votes $V$ of the input capsules are fixed during the EM routing process.

% {\color{red} Dou: Again Eqn.~\ref{dou2} could be thought of as we first do a nonlinear transformation $g$ to extract feature so that ${ {\bf H}^1_{new}, \dots, {\bf H}^L_{new} = g(\bf H}^1, \dots, {\bf H}^L) $. Then we could write $V_{l \rightarrow n}={\bf W}_{l \rightarrow n} {\bf H}^l_{new}$ and $ A^{H_{new}}_l = sigmoid(W^{H_{new}}_l{\bf H}^l_{new})$.}

\paragraph{\em M-Step} for each Gaussian associated with ${\bf \Omega}_n$ consists of finding the mean $\mu_n$ of the votes from input capsules and the variance ${\sigma_n}$ about that mean {for each dimension $h$}:
\begin{eqnarray}
    \mu_n^h &=&\frac{\sum_l C_{l \rightarrow n} V_{l \rightarrow n}^h}{\sum_l C_{l \rightarrow n}} \label{eqn:mu} \\
    (\sigma_n^h)^2 &=& \frac{\sum_l C_{l \rightarrow n} (V_{l \rightarrow n}^h-\mu_n^h)^2}{\sum_l C_{l \rightarrow n}}  \label{eqn:sigma}
\end{eqnarray}
The incremental cost of using an active capsule ${\bf \Omega}_n$ is
\begin{eqnarray}
    cost_n^h = \big(\log(\sigma_n^h) + \frac{1+\log(2\pi)}{2}\big)\sum_l C_{l \rightarrow n}
\end{eqnarray}
% The main difference is that in this work the output of capsule is a scalar {\color{red} Dou: it could be a vector.}, thus we skip the dimension $h$ of capsule output.
The activation probability of capsule ${\bf \Omega}_n$ is calculated by 
%using the minimum description length principle by {\color{red} (need more intuition here)}
\begin{equation}
    A^{\Omega}_n = logistic\big(\lambda(\beta_A - \beta_{\mu} \sum_l C_{l \rightarrow n} - \sum_h {cost}_n^h)\big) \label{eqn:activation}
\end{equation}
where $\beta_A$ is a fixed cost for coding the mean and variance of ${\bf \Omega}_n$ when activating it, $\beta_{\mu}$ is another fixed cost per input capsule when not activating it, and $\lambda$ is an inverse temperature parameter set with a fixed schedule. 
We refer the readers to~\cite{Hinton:2018:ICLR} for more details. 
% $\beta_A$ and $\beta_{\mu}$ are learned discriminatively.

%\begin{eqnarray}
%    \mu_n^h &=&\frac{\sum_l C_{l \rightarrow n} V_{l \rightarrow n}^h}{\sum_i C_{l \rightarrow n}} \label{eqn:mu} \\
%    (\sigma_n^h)^2 &=& \frac{\sum_l C_{l \rightarrow n} (V_{l \rightarrow n}^h-\mu_n^h)^2}{\sum_l C_{l \rightarrow n}}  \label{eqn:sigma}
%\end{eqnarray}
%
%\begin{eqnarray}
%    cost_n^h &=& (\beta_u + \log (\sigma_n^h))\sum_l C_{l \rightarrow n} \\
%    A^{\Omega}_n &=& logistic(\lambda(\beta_a - \sum_h {cost}_n^h)) \label{eqn:activation}
%\end{eqnarray}

\paragraph{\em E-Step} adjusts the assignment probabilities $C_{l \rightarrow *}$ for each input ${\bf \widehat{H}}^l$. 
First, we compute the negative log probability density of the vote $V_{l \rightarrow n}$ from ${\bf \widehat{H}}^l$ under the Gaussian distribution fitted by the output capsule ${\bf \Omega}_n$ it gets assigned to:
\begin{equation}
    p_n = \frac{1}{\sqrt{ 2\pi(\sigma_n)^2}} \exp(- \frac{(V_{l \rightarrow n} - \mu_n)^2}{2(\sigma_n)^2})
\end{equation}
Accordingly, the assignment probability is re-normalized by
\begin{equation}
    C_{l \rightarrow n} = \frac{A^{\Omega}_n p_n}{\sum_{n'} A^{\Omega}_{n'} p_{n'}} \label{eqn:assignment}
\end{equation}

% \paragraph{\color{red} \em Discussion: advantages of EM.} 

As has been stated above, EM routing is a more powerful routing algorithm, which can better estimate the agreement by allowing active capsules to receive a cluster of similar votes. In addition, it assigns an additional activation probability $A$ to represent the probability of whether each capsule is present, rather than the length of vector.

\section{Experiment}

\begin{table*}[t]
  \centering
  \begin{tabular}{c|l||r c c||c|c}
    {\bf \#}    &   {\bf Model} &  \bf {\# Para.} & \bf {Train}  &   \bf Decode  &    \bf  BLEU  &   $\bigtriangleup$ \\
    \hline
    1   &   \textsc{Transformer-Base}      & 88.0M &   1.79  & 1.43 &  27.31   &   --\\
    2   &   ~~ + Linear Combination~\cite{Dou:2018:EMNLP}         &   +14.7M    &   1.57  &   1.36  &  27.73   &   +0.42\\
    \hline
    3   &   ~~ + Dynamic Combination             & +25.2M   & 1.50  & 1.30 & 28.33  &   +1.02\\
    \hdashline
    4   &   ~~ + Dynamic Routing        & +37.8M & 1.37  & 1.24 & 28.22 &   +0.91\\
    5   &   ~~ + EM Routing       & +56.8M & 1.10  & 1.15 & 28.81 &   +1.50\\
  \end{tabular}
  \caption{Translation performance on WMT14 English$\Rightarrow$German translation task. ``\# Para.'' denotes the number of parameters, and ``Train'' and ``Decode'' respectively denote the training (steps/second) and decoding (sentences/second) speeds.} % on Tesla P40.}
  \label{tab:main}
\end{table*}

\begin{table*}[t]
  \centering
  \begin{tabular}{l|l||rl|rl}%|c}
    \multirow{2}{*}{\bf System}  &   \multirow{2}{*}{\bf Architecture}  &  \multicolumn{2}{c|}{\bf En$\Rightarrow$De}  & \multicolumn{2}{c}{\bf Zh$\Rightarrow$En}\\
    \cline{3-6}
        &   &   \# Para.    &   BLEU    &   \# Para.    &   BLEU\\
    \hline \hline
    \multicolumn{6}{c}{{\em Existing NMT systems}} \\
    \hline
    \cite{wu2016google} &   \textsc{Rnn} with 8 layers           &   N/A &   26.30  &  N/A &  N/A  \\ 
    \cite{Gehring:2017:ICML}  &   \textsc{Cnn} with 15 layers  &   N/A &   26.36  &  N/A &  N/A  \\
    \hline
    \multirow{2}{*}{\cite{Vaswani:2017:NIPS}} &   \textsc{Transformer-Base}    &  65M &   27.3  &    N/A & N/A \\ 
    &  \textsc{Transformer-Big}               &  213M &  28.4  &  N/A  &  N/A\\ 
    \hdashline
    \cite{hassan2018achieving}  &   \textsc{Transformer-Big}  &  N/A  & N/A &  N/A  &  24.2\\
    %\textsc{RNNSearch} + Ensemble \cite{wang2017sogou} & 22.9 (dev) & - \\ 
    \hline\hline
    \multicolumn{6}{c}{{\em Our NMT systems}}   \\ \hline
    \multirow{4}{*}{\em this work}  &   \textsc{Transformer-Base}  &  88M  &  {{27.31}}   &    108M  & 24.13\\
    &   ~~ + EM Routing   &    123M  & {{28.81$^{\dag}$}}   &    143M &   24.81$^{\dag}$\\
    \cline{2-6}
    &   \textsc{Transformer-Big}    & 264M  &   28.58      &  304M    & 24.56\\
    &   ~~ + EM Routing   &  490M & {{28.97$^{\dag}$}}
       &  530M &    25.00$^{\dag}$\\
  \end{tabular}
    \caption{Comparing with existing NMT systems on WMT14 English$\Rightarrow$German (``En$\Rightarrow$De'') and WMT17 Chinese$\Rightarrow$English (``Zh$\Rightarrow$En'') tasks. 
    %``+ Deep Representations'' denotes ``+ Hierarchical Aggregation + $\mathcal{L}_{diversity}$''. 
    ``$\dag$'' indicates statistically significant difference ($p < 0.01$) from the \textsc{Transformer} baseline.}
    \label{tab:exist}
  \end{table*}

\subsection{Setting}

% \paragraph{Dataset.}
We conducted experiments on two widely-used WMT14 English $\Rightarrow$ German (En$\Rightarrow$De) and WMT17 Chinese $\Rightarrow$ English (Zh$\Rightarrow$En) translation tasks and compared our model with results reported by previous work~\cite{Gehring:2017:ICML,Vaswani:2017:NIPS,hassan2018achieving}.
For the En$\Rightarrow$De task, the training corpus consists of about $4.56$ million sentence pairs.  We used newstest2013 as the development set and newstest2014 as the test set.
For the Zh$\Rightarrow$En task, we used all of the available parallel data, consisting of about $20$ million sentence pairs. We used newsdev2017 as the development set and newstest2017 as the test set.
All the data had been tokenized and segmented into subword symbols using byte-pair encoding with 32K merge operations \cite{sennrich2016neural}.
% We learned a BPE model with 32K merge operations for both datasets. 
%The Chinese data had been tokenized using {\it Jieba}~\footnote{https://github.com/fxshy/jieba}. English sentences were tokenized using the scripts provided in Moses. 
%The Zh$\Rightarrow$En corpus consists of 20M sentence pairs, and the En$\Rightarrow$De corpus consists of 4M sentence pairs. We followed previous work to select the validation and test sets.
%Byte-pair encoding (BPE) was employed to alleviate the Out-of-Vocabulary problem \cite{sennrich2016neural} with 32K merge operations for both language pairs.
We used 4-gram NIST BLEU score \cite{papineni2002bleu} as the evaluation metric, and {\em sign-test} \cite{Collins05} for statistical significance test.

% \paragraph{Models.}
We evaluated the proposed approaches on the Transformer model~\cite{Vaswani:2017:NIPS}.
We followed the configurations in~\cite{Vaswani:2017:NIPS}, and reproduced their reported results on the En$\Rightarrow$De task.
The parameters of the proposed models were initialized by the pre-trained model.
% We have tested both \emph{Base} and \emph{Big} models, which differ at hidden size (512 vs. 1024), filter size (2048 vs. 4096) and the number of attention heads (8 vs. 16). \footnote{Here ``filter size'' refers to the hidden size of the feed-forward network in the Transformer model.}
All the models were trained on eight NVIDIA P40 GPUs where each was allocated with a batch size of 4096 tokens.
In consideration of computation cost, we studied model variations with \textsc{Transformer-Base} model on En$\Rightarrow$De task, and evaluated overall performance with {\textsc{Transformer-Base} and} \textsc{Transformer-Big} model on both Zh$\Rightarrow$En and En$\Rightarrow$De tasks.

\subsection{Results}

\subsubsection{Model Variations}
Table \ref{tab:main} shows the results on WMT14 En$\Rightarrow$De translation task. 
As one would expect, the linear combination (Row 2) improves translation performance by +0.42 BLEU points, indicating the necessity of aggregating layers for deep NMT models.

\iffalse
As we could see, the proposed approaches outperform both the baseline model and the linear combination in all cases, % improve the translation quality in all cases, 
although there are still considerable differences among different variations.
\fi

% \paragraph{Model Complexity}

% \paragraph{Dynamic Layer Aggregation} (Rows 3-5):
All dynamic aggregation models (Rows 3-5) consistently outperform its static counterpart (Row 2), demonstrating the superiority of the dynamic mechanisms. Among the model variations, the simplest strategy -- dynamic combination (Row 3) surprisingly improves performance over the baseline model by up to +1.02 BLEU points. 
Benefiting from the advanced routing-by-agreement algorithm, the dynamic routing strategy can achieve similar improvement. The EM routing further improves performance by better estimating the agreement during the routing. These findings suggest potential applicability of capsule networks to natural language processing tasks, which has not been fully investigated yet.

All the dynamic aggregation strategies introduce new parameters, ranging from 25.2M to 56.8M. Accordingly, the training speed would decrease due to more efforts to train the new parameters.
Dynamic aggregation mechanisms only marginally decrease decoding speed, with EM routing being the slowest one, which decreases decoding speed by 19.6\%.

% Unless otherwise stated, all the following experiments used ``EM routing''.

\iffalse
Even the simplest dynamic routing strategy, namely dynamic combination (row 3), could achieve surprisingly good results compared to its static counterpart, which is up to +1.06 BLEU points better than the baseline model. With sophisticated routing strategy, dynamic routing (row 4) and EM routing (row 5) could enhance the performance of the baseline model and significantly outperform baseline by +0.91 and +1.50 BLEU points. These findings have shown that capsule networks may have promising applications in natural language processing, which has not been fully investigated yet.
\fi

\subsubsection{Main Results}
Table~\ref{tab:exist} lists the results on both WMT17 Zh$\Rightarrow$En and WMT14 En$\Rightarrow$De translation tasks. As seen, dynamically aggregating layers consistently improves translation performance across NMT models and language pairs, which demonstrating the effectiveness and universality of the proposed approach.
It is worth mentioning that \textsc{Transformer-Base} with EM routing outperforms the vanilla \textsc{Transformer-Big} model, with only less than half of the parameters, demonstrating our model could utilize the parameters more efficiently and effectively.

\subsection{Analysis of Iterative Routing}

We conducted extensive analysis from different perspectives to better understand the iterative routing process. %in terms of its effect on different NMT components, impact of different intermediate function choices, as well as building the ability of handling long sentences.
All results are reported on the development set of En$\Rightarrow$De task with ``\textsc{Transformer-Base} + EM routing'' model.
\begin{figure}[h]
\centering
\includegraphics[width=0.47\textwidth]{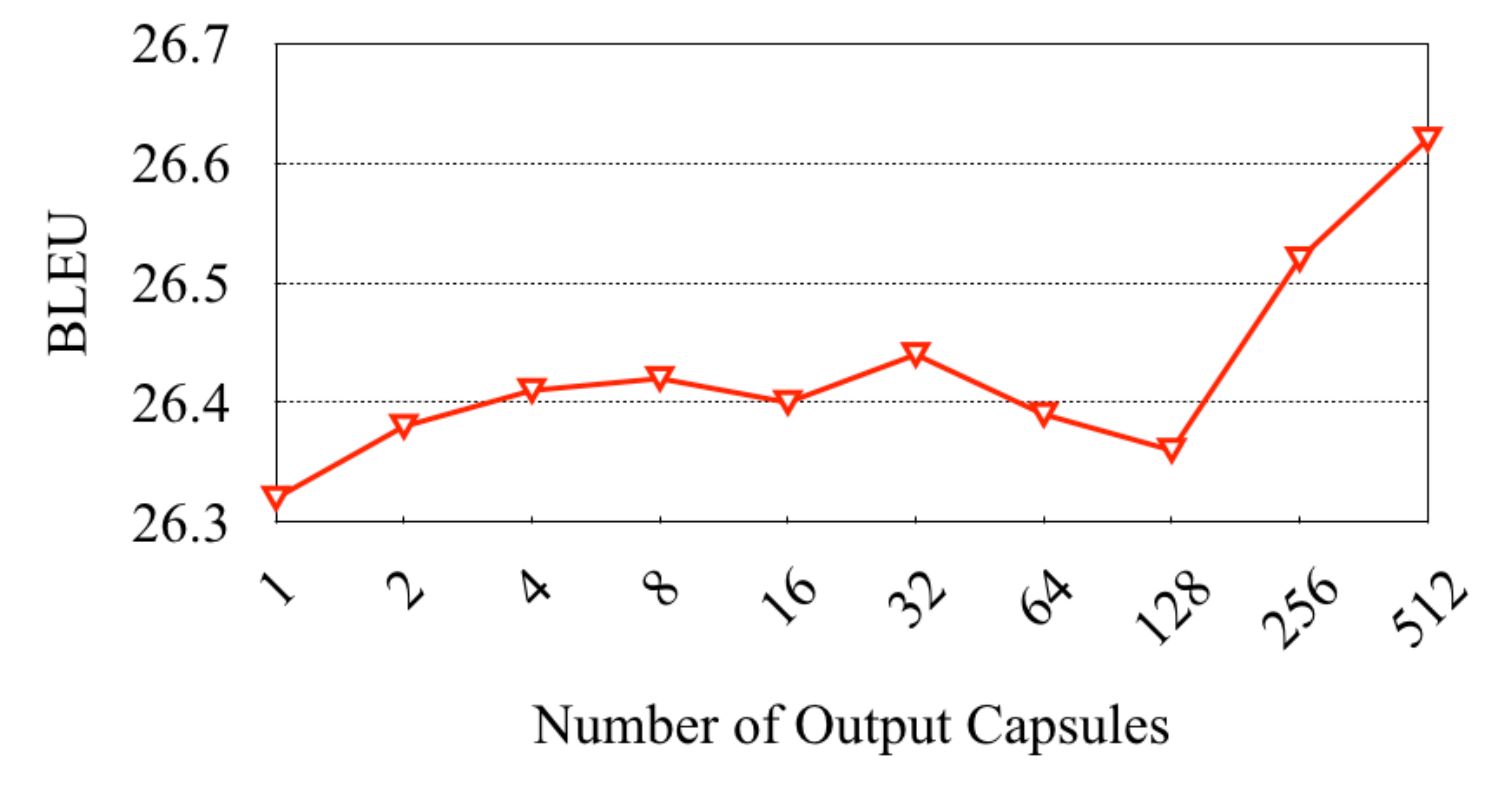}
\caption{Impact of number of output capsules.}
\label{fig:output_capsule_number}
\end{figure}
% In the first series of experiments, we evaluated the impact of different components of iterative routing on translation quality.

%\vspace{-10pt}
\subsubsection{Impact of the Number of Output Capsules}
The number of output capsules $N$ is a key parameter for our model, as shown in Figure~\ref{fig:dynamic}. 
We plot in Figure~\ref{fig:output_capsule_number} the BLEU score with different number of output capsules.
Generally, the BLEU score goes up with the increase of the capsule numbers.
As aforementioned, to make the dimensionality of the final output be consistent with hidden layer (i.e. $d$), the dimensionality of each capsule output is $d/N$. 
When $n$ increases, the dimensionality of capsule output $d/N$ decreases (the minimum value is 1), which may lead to more subtle representations of different properties of the input.

% We have tried different values to test the effect of the number of output capsules and the results are plotted in Figure~\ref{fig:capsule_number}. 
% As seen, our model could achieve the highest BLEU points on the development set when the output capsules is maximized, {\it i.e.} equals to the hidden size of the Transformer, which is not so surprising given this setting has the most parameters. However, the performance does not drop significantly as we reduce the number of output capsules, which could demonstrate the robustness of our model.

\begin{figure}[h]
\centering
\includegraphics[width=0.47\textwidth]{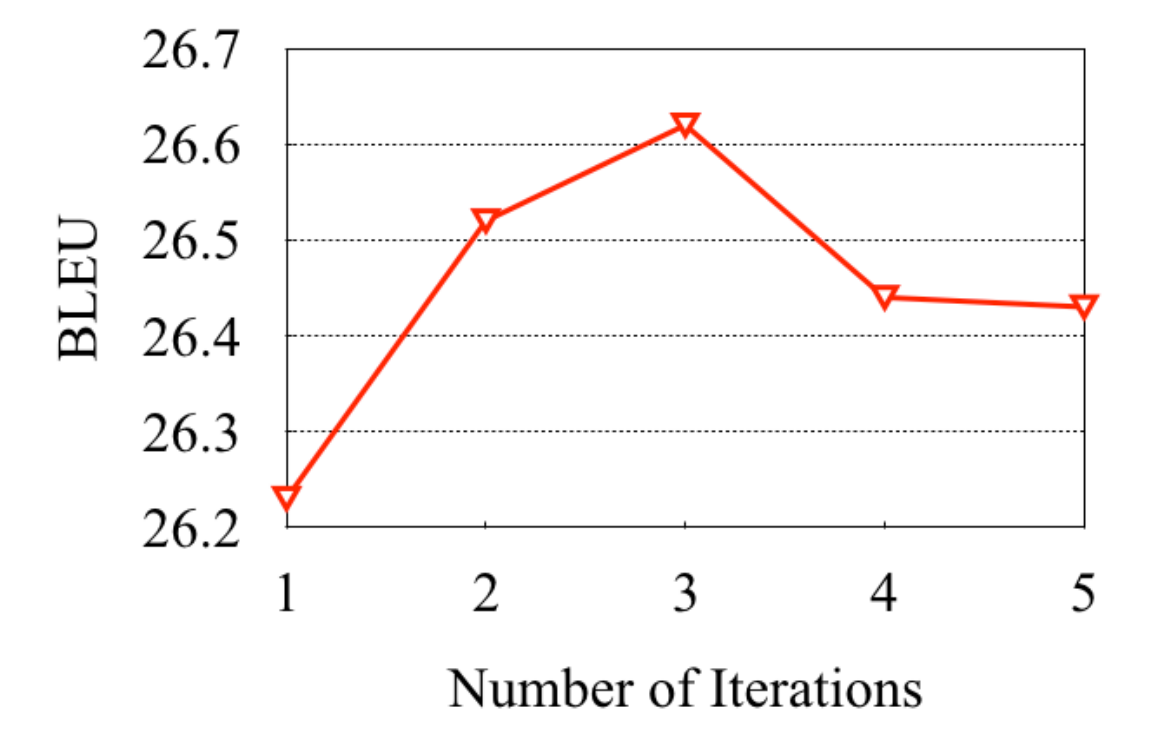}
\caption{Impact of routing iterations.}
\label{fig:iter}
\end{figure}
%\vspace{-10pt}
\subsubsection{Impact of Routing Iterations}
Another key parameter is the iteration of the iterative routing $T$, which affects the estimation of the agreement.  
As shown in Figure~\ref{fig:iter}, the BLEU score typically goes up with the increase of the iterations $T$, while the trend does not hold when $T > 3$. 
This indicates that more iterations may over-estimate the agreement between two capsules, thus harms the performance.
The optimal iteration $3$ is also consistent with the findings in previous work~\cite{Sabour:2017:NIPS,Hinton:2018:ICLR}.

\begin{table}[h]
  \centering
  \begin{tabular}{c|c||c}
    {\bf Model}   &      \bf {Construct ${\bf \widehat{H}}^l$ with}     &   {\bf BLEU}\\  
    \hline
    \textsc{Base}    &   N/A      & 25.84\\
    \hline
    \hline
    %\hline
    \multirow{2}{*}{\textsc{Ours}}		 
 		&    $F_l({\bf H}^l)$ & 26.18\\
 		&    $F_l({\bf H}^1, \dots, {\bf H}^L)$ & 26.62\\
  \end{tabular}
  \caption{Impact of functions to construct input capsules.}
  \label{tab:extracting}
\end{table}

%\vspace{-10pt}
\subsubsection{Impact of Functions to Construct Input Capsules}
For the iterative routing models, we use ${\bf \widehat{H}}^l = F_l({\bf H}^1, \dots, {\bf H}^L)$ instead of ${\bf \widehat{H}}^l = F_l({\bf H}^l)$ to construct each input capsule ${\bf \widehat{H}}^l$. 
% This step can be regarded as a feature extraction function to first extract features from the concatenation of the original layer representations. 
Table~\ref{tab:extracting} lists the comparison results, which shows that the former indeed outperforms the latter.
We attribute this to that $F_l({\bf H}^1, \dots, {\bf H}^L)$ is more representative by extracting features from the concatenation of the original layer representations.

\iffalse
\begin{figure}[h]
\centering
\subfloat[$1^{st}$ Iter.]{\includegraphics[width=0.12\textwidth]{figures/matrix/num_4_iter1.png}} \hfill
\subfloat[$2^{nd}$ Iter.]{\includegraphics[width=0.12\textwidth]{figures/matrix/num_4_iter2.png}} \hfill
\subfloat[$3^{rd}$ Iter.]{\includegraphics[width=0.12\textwidth]{figures/matrix/num_4_iter3.png}}
\caption{Agreement distribution with 4 output capsules (x-axis), and y-axis is 6 input capsules.} % and x-axis is 4 output capsules.}
\end{figure}
\fi

\iffalse
\begin{figure}[h]
\centering
\subfloat[$1^{st}$ Iteration]{\includegraphics[width=0.47\textwidth]{figures/512_routing_iter1.pdf}} \\
\subfloat[$2^{nd}$ Iteration]{\includegraphics[width=0.47\textwidth]{figures/512_routing_iter2.pdf}} \\
\subfloat[$3^{rd}$ Iteration]{\includegraphics[width=0.47\textwidth]{figures/512_routing_iter3.pdf}}
\caption{Agreement distribution with 6 input capsules (y-axis) and 512 output capsules (x-axis). Darker color denotes higher agreement.}
\label{fig:agreement}
\end{figure}
\fi

\begin{figure}[h]
\centering
\includegraphics[width=0.47\textwidth]{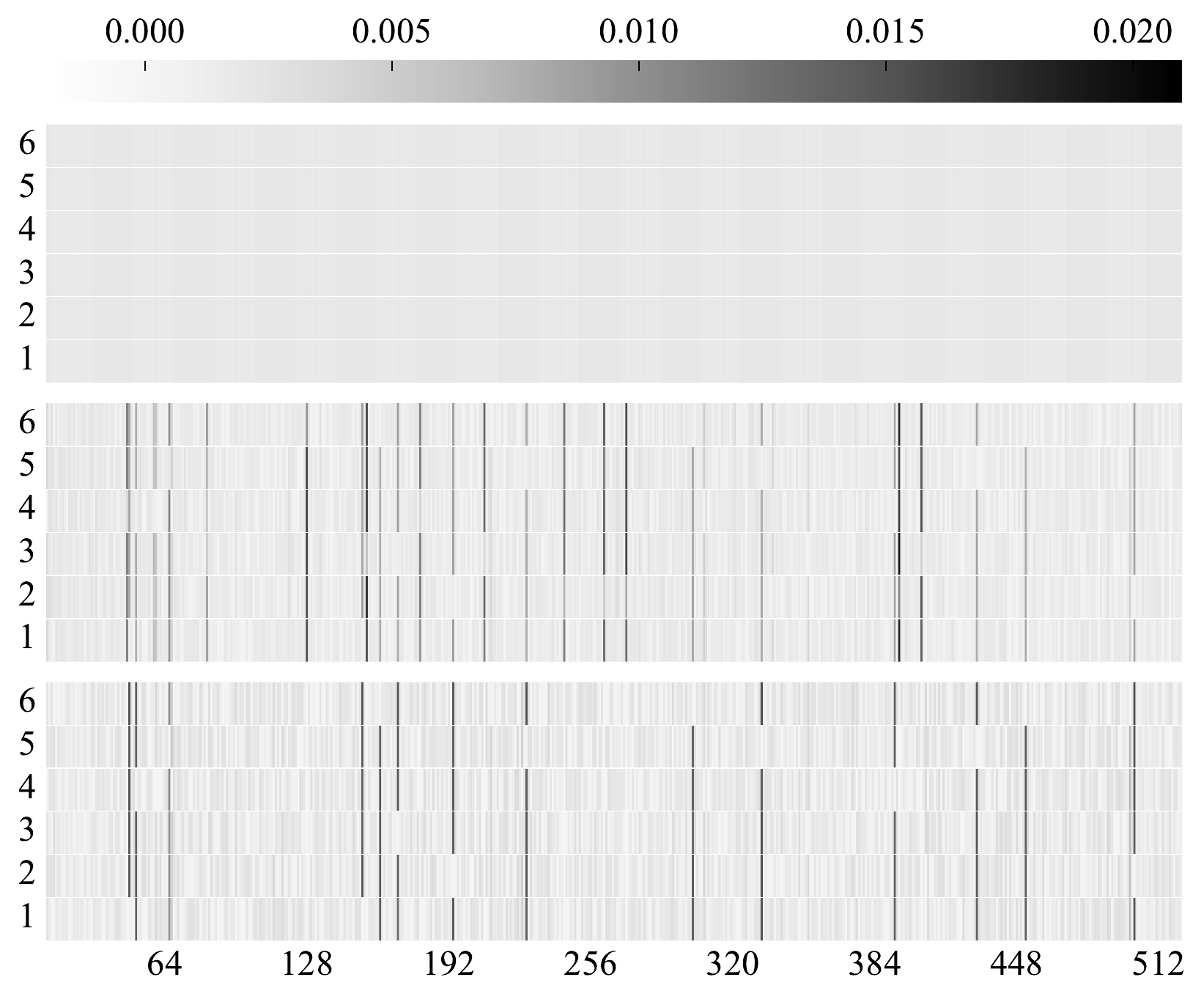}
\caption{Agreement distribution with 6 input capsules (y-axis) and 512 output capsules (x-axis). Darker color denotes higher agreement. The three heatmaps from top to bottom are respectively the $1^{st}$ to $3^{rd}$ iterations.}
% 512 Routing Patterns. Diversity (calculate by rows by columns are the same: 52863, 29083, 57172); Entropy (6.69, 3.40, 5.80).}
\label{fig:agreement}
\end{figure}

\subsubsection{Visualization of Agreement Distribution}
The assignment probability $C_{l \rightarrow n}$ {before M step} with $\sum_n C_{l \rightarrow n}=1$ denotes the agreement between the input capsule ${\bf \widehat{H}}^l$ and the output capsule ${\bf \Omega}_n$, which is determined by the iterative routing. 
A higher agreement $C_{l \rightarrow n}$ denotes that the input capsule ${\bf \widehat{H}}^l$ prefers to send its representation to the output capsule ${\bf \Omega}_n$. 
We plot in Figure~\ref{fig:agreement} the alignment distribution in different routing iterations. 
In the first iteration (top panel), the initialized uniform distribution is employed as the agreement distribution, and each output capsule equally attends to all the input capsules.
As the iterative routing goes, the input capsules learns to send their representations to proper output capsules, and accordingly output capsules are more likely to capture distinct features. 
We empirically validate our claim from the following two perspectives.
% (more darker cells in the later iterations). 
% by capturing the most active features from the input capsules. These results provide support for the claim that benefiting from the high-dimensional coincidence filtering, capsule neurons are able to ignore all but the most active feature from the input capsules.

We use the entropy to measure the skewness of the agreement distributions:
\begin{equation}
   entropy = \frac{1}{N \cdot L} \sum_{n=1}^N \sum_{l=1}^L C_{l \rightarrow n} \log C_{l \rightarrow n}
\end{equation}
A lower entropy denotes a more skewed distribution, which indicates that the input {capsules are more certain about which output capsules should be routed more information. }The entropies of the three iterations are respectively 6.24, 5.93, 5.86, which indeed decreases as expected.

To validate the claim that different output capsules focus on different subsets of input capsules, we measure the diversity between each two output capsules. Let ${\bf C}_n = \{C_{1 \rightarrow n}, \dots, C_{L \rightarrow n}\}$ be the agreement probabilities assigned to the output capsule ${\bf \Omega}_n$, we calculate the diversity among all the output capsules as
\begin{equation}
   diversity = \frac{1}{\binom{N}{2}} \sum_{i=1}^N \sum_{j=i+1}^N  \cos({\bf C}_i, {\bf C}_j)
\end{equation}
%where $\binom{N}{2}$ is the %binomial coefficient, which serves as the normalizing factor. 
A higher diversity score denotes that output capsules attend to different subsets of input capsules.
The diversity scores of the three iterations are respectively 0.0, 0.09, and 0.18, which reconfirm our observations.

\subsection{Effect on Encoder and Decoder}

\begin{table}[h]
  \centering
  \begin{tabular}{c|c|c||c}
    \multirow{2}{*}{\bf Model}   & \multicolumn{2}{c||}{\bf Applied to}      &   \multirow{2}{*}{\bf BLEU}\\  
    \cline{2-3}
                    &       \em Encoder &   \em Decoder   &   \\
    \hline
    \textsc{Base}    &   N/A  &   N/A  &    25.84\\
    \hline
    \hline
    \multirow{3}{*}{\textsc{Ours}}		&   \checkmark  &   \texttimes   & 26.33 \\
    		                            &	\texttimes  &	\checkmark   &   26.34\\
     		                            &   \checkmark  &   \checkmark   & \bf 26.62 \\
  \end{tabular}
  \caption{Effect of EM routing on encoder and decoder.}
  \label{tab:component}
\end{table}

Both encoder and decoder are composed of a stack of $L$ layers, which may benefit from the proposed approach. In this experiment, we investigate how our model affects the two components, as shown in Table~\ref{tab:component}. Aggregating layers of encoder or decoder individually consistently outperforms the vanilla baseline model, and exploiting both components further improves performance. These results provide support for the claim that aggregating layers is useful for both understanding input sequence and generating output sequence.

\subsection{Length Analysis}

\begin{figure}[h]
\centering
\includegraphics[height=0.28\textwidth]{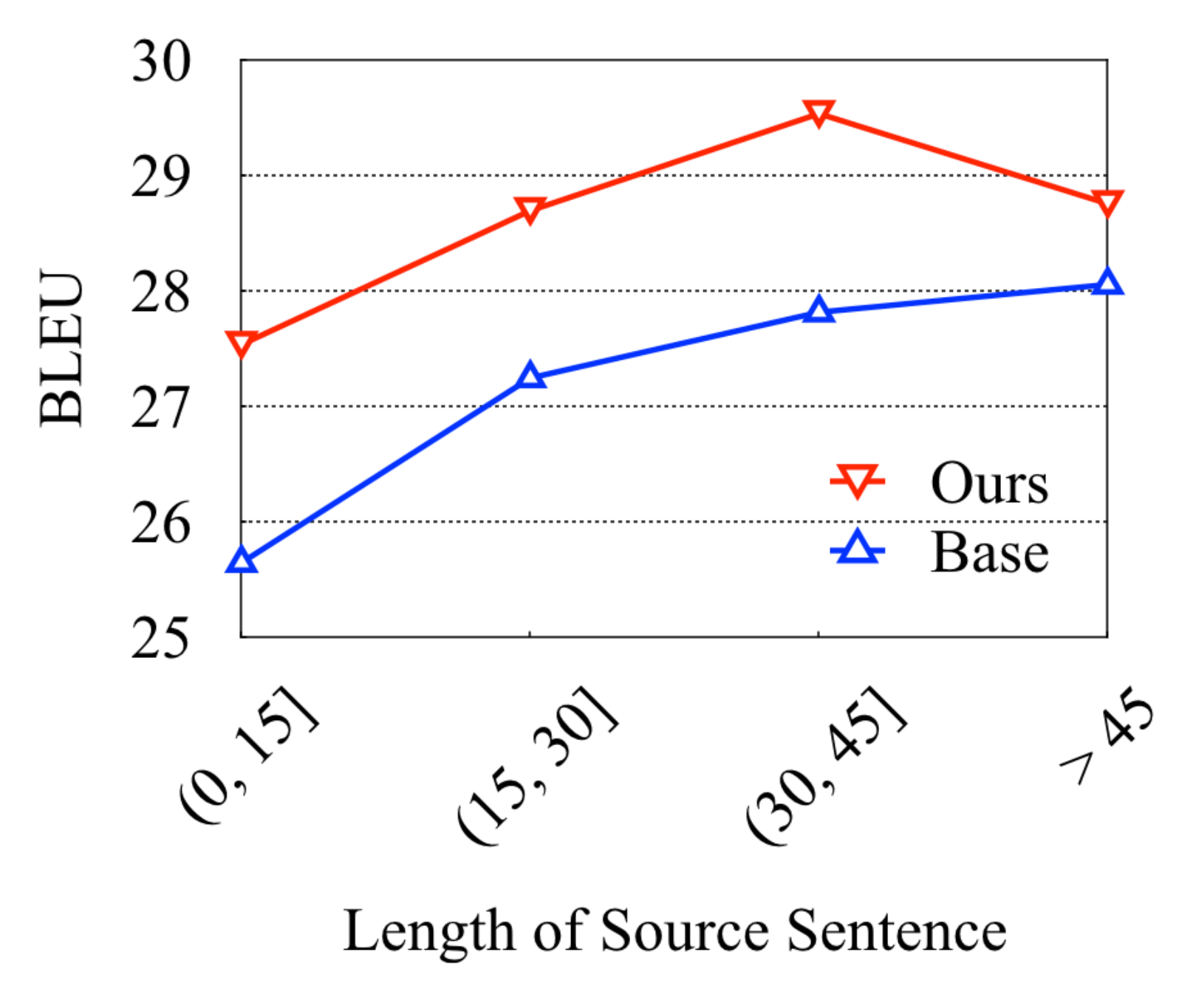}
\caption{BLEU scores on the En$\Rightarrow$De test set with respect to various input sentence lengths.}
\label{fig:length}
\end{figure}

Following~\citeauthor{Bahdanau:2015:ICLR} \shortcite{Bahdanau:2015:ICLR} and ~\citeauthor{tu2016modeling} \shortcite{tu2016modeling},
we grouped sentences of similar lengths together and computed the BLEU score for each group, as shown in Figure~\ref{fig:length}.
Generally, the performance of \textsc{Transformer-Base} goes up with the increase of input sentence lengths. We attribute this to the strength of self-attention mechanism to model global dependencies without regard to their distance.
Clearly, the proposed approaches outperform the baseline in all length segments. 
% The sentences of length {less than or equal to 45} seem to benefit from our models the most, which may not be surprising given we discarded the sentences whose length are greater than 50 during training. We believe incorporating more long sentences into the training could further improve the performance of our model on long segments.

\iffalse

\begin{figure*}[h]
\centering
\includegraphics[width=1.0\textwidth]{figures/4_routing.jpeg}
\caption{4 Routing Patterns.}
\label{fig:4}

\end{figure*}
\begin{figure*}[h]
\centering
\includegraphics[width=1.0\textwidth]{figures/32_routing.jpeg}
\caption{32 Routing Patterns.Diversity: (397, 613, 715). Entropy (13.03, 2.88, 3.08).}
\label{fig:32}
\end{figure*}

\fi

\section{Related Work}

% The two main sub-field related to our work are deep representation learning and dynamic aggregation.
Our work is inspired by research in the field of {\em exploiting deep representation} and {\em capsule networks}.

\paragraph{Exploiting Deep Representation}
Exploiting deep representations have been studied by various communities, from computer vision to natural language processing. 
\citeauthor{he2016deep} \shortcite{he2016deep} propose a residual learning framework, combining layers and encouraging gradient flow by simple short-cut connections. \citeauthor{Huang:2017:CVPR} \shortcite{Huang:2017:CVPR} extend the idea by introducing densely connected layers which could better strengthen feature propagation and encourage feature reuse. Deep layer aggregation \cite{Yu:2018:CVPR} designs architecture to fuse information iteratively and hierarchically.

Concerning natural language processing, \citeauthor{Peters:2018:NAACL} \shortcite{Peters:2018:NAACL}
 have found that combining different layers is helpful and their model significantly improves state-of-the-art models on various tasks. 
 Researchers have also explored fusing information for NMT models and demonstrate aggregating layers is also useful for NMT ~\cite{shen2018dense,wang2018multi,Dou:2018:EMNLP}.
However, all of these works mainly focus on static aggregation in that their aggregation strategy is independent of specific hidden states. In response to this problem, we introduce  dynamic principles into layer aggregation. In addition, their approaches are a fixed policy without considering the representation of the final output, while the routing-by-agreement mechanisms are able to aggregate information according to the final representation.

 \paragraph{Capsule Networks}
 The idea of dynamic routing is first proposed by~\citeauthor{Sabour:2017:NIPS}~\shortcite{Sabour:2017:NIPS}, which aims at addressing the representational limitations of convolutional and recurrent neural networks for image classification. 
%The proposed capsule networks mainly replace max-pooling with routing-by-agreement, which is referred to as ``dynamic routing'' in this paper. 
The iterative routing procedure is further improved by using Expectation-Maximization algorithm to better estimate the agreement between capsules~\cite{Hinton:2018:ICLR}. 
% {\color{red} a few citations on CV tasks}
In computer vision community, 
\citeauthor{xi2017capsule}
\shortcite{xi2017capsule} explore its application on CIFAR data with higher dimensionality. \citeauthor{lalonde2018capsules}
\shortcite{lalonde2018capsules} apply capsule networks on object segmentation task.

The applications of capsule networks in natural language processing tasks, however, have not been widely investigated to date. \citeauthor{zhao2018investigating} \shortcite{zhao2018investigating} testify capsule networks on text classification tasks and \citeauthor{Gong:2018:arXiv} \shortcite{Gong:2018:arXiv} propose to aggregate a sequence of vectors via dynamic routing for sequence encoding. 
To the best of our knowledge, this work is the first to apply the idea of dynamic routing to NMT.
% no one has applied the idea of dynamic routing in machine translation so far.

\section{Conclusion}
In this work, we propose several methods to dynamically aggregate layers for deep NMT models. 
Our best model, which utilizes EM-based iterative routing to estimate the agreement between inputs and outputs, has achieved significant improvements over the baseline model across language pairs. 
By visualizing the routing process, we find that capsule networks are able to extract most active features shared by different inputs.
Our study suggests potential applicability of capsule networks across computer vision and natural language processing tasks for aggregating information of multiple inputs.

% Future directions include validating our approach on other NMT architectures~\cite{chen2018the,Gehring:2017:ICML} as well as other natural language processing tasks {\color{red}such as text generation and abstract summarization}.

Future directions include validating our approach on other NMT architectures such as RNN~\cite{chen2018the} and CNN~\cite{Gehring:2017:ICML}, as well as on other NLP tasks such as dialogue and reading comprehension.
It is also interesting to combine with other techniques~\cite{shaw2018self,Li:2018:EMNLP,Dou:2018:EMNLP,Yang:2018:EMNLP,Yang:2019:AAAI,Kong:2019:AAAI} to further boost the performance of Transformer.

% Experimental results on WMT14 English$\Rightarrow$German and WMT17 Chinese$\Rightarrow$English show that our model could consistently outperform the state-of-the-art Transformer model by 0.68 and 0.44, respectively. By visualizing the routing process, we find that there are distinctions between output capsules.

% We would like to test our models on other architectures like RNN~\cite{chen2018the} or CNN~\cite{Gehring:2017:ICML} in the future, as well as come up with more sophisticated routing algorithm.

\balance
\bibliography{main}
\bibliographystyle{aaai}
\end{document}